\documentclass[anon]{l4dc2026}

\usepackage{times}
\usepackage{latexsym}
\usepackage{amssymb}
\usepackage{booktabs}
\usepackage{enumitem}
\usepackage{color}
\usepackage{upgreek}
\usepackage{helvet} 
\usepackage{courier} 
\usepackage{subcaption}
\usepackage{multirow}
\usepackage{wrapfig}

\newcounter{subfig}[figure] 



\newcommand{\context}[1]{}

\newcommand{\algo}[0]{\textsc{tiger}}

\title[TIGER]{\algo-\textsc{marl}: Enhancing Multi-Agent Reinforcement Learning with \textbf{\underline{T}}emporal \textbf{\underline{I}}nformation through \textbf{\underline{G}}raph-based \textbf{\underline{E}}mbeddings and \textbf{\underline{R}}epresentations }

\author{%
\Name{Nikunj Gupta}$^{1}$\thanks{Equal contribution.} \Email{nikunj@usc.edu}\\
\Name{Ludwika Twardecka}$^{1}$\footnotemark[1] \Email{twardeck@usc.edu}\\
\Name{James Zachary Hare}$^{2}$ \Email{james.z.hare.civ@army.mil}\\
\Name{Jesse Milzman}$^{2}$ \Email{jesse.m.milzman.civ@army.mil}\\
\Name{Rajgopal Kannan}$^{2}$ \Email{rajgopal.kannan.civ@army.mil}\\
\Name{Viktor Prasanna}$^{1}$ \Email{prasanna@usc.edu}\\
\addr $^{1}$University of Southern California\\ 
\addr $^{2}$DEVCOM Army Research Office
}

\begin{document}

\maketitle

\begin{abstract}
In this paper, we propose capturing and utilizing \textit{Temporal Information through Graph-based Embeddings and Representations} or \textbf{\algo} to enhance multi-agent reinforcement learning (MARL). We explicitly model how inter-agent coordination structures evolve over time. While most MARL approaches rely on static or per-step relational graphs, they overlook the temporal evolution of interactions that naturally arise as agents adapt, move, or reorganize cooperation strategies. Capturing such evolving dependencies is key to achieving robust and adaptive coordination. To this end, \algo\ constructs dynamic temporal graphs of MARL agents, connecting their current and historical interactions. It then employs a temporal attention-based encoder to aggregate information across these structural and temporal neighborhoods, yielding time-aware agent embeddings that guide cooperative policy learning. Through extensive experiments on two coordination-intensive benchmarks, we show that \algo\ consistently outperforms diverse value-decomposition and graph-based MARL baselines in task performance and sample efficiency. Furthermore, we conduct comprehensive ablation studies to isolate the impact of key design parameters in \algo, revealing how structural and temporal factors can jointly shape effective policy learning in MARL. 
All codes can be found here: \url{https://github.com/Nikunj-Gupta/tiger-marl}. 
\end{abstract}


\section{Introduction} 
\label{sec:introduction} 

\context{Challenges in MARL} 

Multi-Agent Reinforcement Learning (MARL) has emerged as a powerful paradigm across diverse domains such as autonomous driving \citep{shalev2016safe, palanisamy2020multi}, resource management \citep{guestrin2001multiagent}, and swarm robotics \citep{orr2023multi, rizk2019cooperative}. Its growing success in cooperative artificial intelligence reflects its ability to model shared decision-making among agents that collectively optimize a long-term objective in complex, interactive systems. As the field continues to advance, capturing inter-agent interactions has become central to achieving adaptive and effective coordination, whether through explicit communication, implicit information exchange, or integrating neighboring agents’ signals into local decision-making. 

\begin{wrapfigure}{r}{0.49\textwidth}
    \centering
    \includegraphics[width=\linewidth]{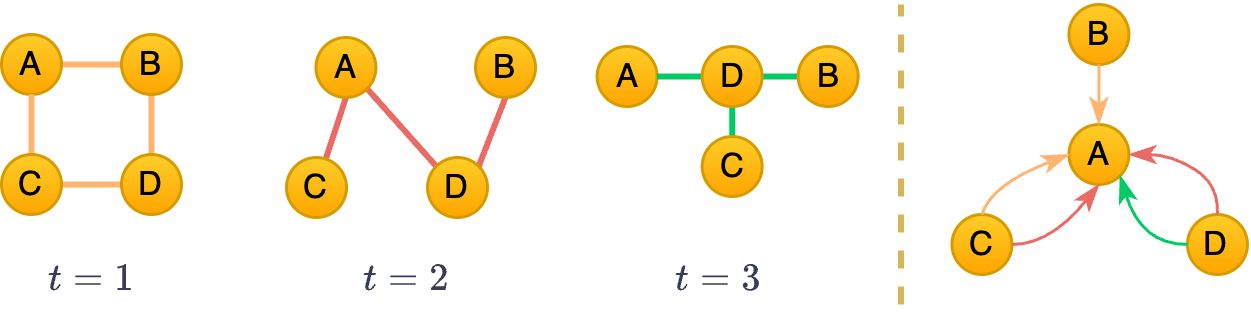}
    \caption{\textbf{Evolving coordination in MARL.} Temporal interaction patterns can reveal useful relational cues to enhance learning.}
    \label{fig:temporal_graph}
\end{wrapfigure}

Irrespective of the application domain, agents in cooperative tasks must effectively reason about their interactions to agree on the best strategy at a given moment. To achieve that, both present and past context play an equally critical role. Fig.~\ref{fig:temporal_graph} illustrates this, showing that as agents progress through time, their connectivity and influence patterns shift. Restricting the model to current inter-agent connections limits its ability to capture how coordination structures evolve as agents adapt their roles and strategies in response to environmental dynamics. Modeling evolving relational dependencies can provide temporal cues that help agents anticipate, adapt, and collaborate more effectively. This becomes especially valuable in partially observable settings, where agents must infer latent coordination structures from limited observations. 

Most MARL approaches either ignore inter-agent relationships or rely on static coordination graphs that capture dependencies only at a single timestep. To overcome this, we draw on advances in graph representation learning. Graph Neural Networks (GNNs) \citep{veličković2018graphattentionnetworks} have been widely used in MARL to model structural dependencies and integrate agent information~\citep{liu2025survey}. Extending to temporal dependencies, Temporal GNNs model how node and edge structures evolve over time, providing a natural way to represent dynamic coordination patterns. These have shown strong performance in domains such as social networks, financial forecasting, and recommendation systems \citep{feng2024comprehensivesurveydynamicgraph, barros2021surveyembeddingdynamicgraphs}. To the best of our knowledge, however, their application to MARL remains largely unexplored. 

In this paper, we propose \algo, a principled approach for modeling evolving agent interaction patterns in MARL. \algo\ introduces a temporal graph construction mechanism that defines each agent’s structural and temporal neighborhoods by explicitly linking current and historical interactions. It then employs an attention-based temporal graph encoder that aggregates information across these neighborhoods to produce time-aware agent embeddings. Through temporal attention, the encoder adaptively balances recent and past interactions, allowing agents to focus on the most relevant dynamics as coordination patterns evolve. \algo\ serves as an expressive temporal reasoning module within MARL, allowing agents to anticipate shifting dependencies and act coherently under partial observability. We demonstrate that this approach substantially improves task performance and sample efficiency over strong value-factorization and graph-based MARL baselines across two diverse coordination-intensive benchmarks. Finally, we conduct detailed ablations to analyze how temporal depth and static connectivity influence coordination, offering insights into the trade-offs that govern effective static-temporal reasoning in MARL.

\section{Related Works} 
\label{sec:related} 

\paragraph{MARL.} Foundational value-factorization methods such as VDN \citep{sunehag2018value} and QMIX \citep{rashid2018qmix} decompose a global action-value function into individual utilities. Extensions including QTRAN \citep{son2019qtran} and QPLEX \citep{wang2021qplex} refined this decomposition to improve representational expressiveness. However, these approaches assume implicit or static coordination, where cooperation arises indirectly through reward structure rather than explicit modeling of agent dependencies. To overcome this limitation, graph-based MARL frameworks represent agents as nodes and their interactions as edges, enabling structured message passing \citep{li2021deepimplicitcoordinationgraphs,bohmer2020deep,duan2024groupawarecoordinationgraphmultiagent,duan2024inferring,gupta2025deep,kok2006collaborative}. DCG \citep{bohmer2020deep} introduced pairwise factorization on fixed graphs, DICG \citep{li2021deepimplicitcoordinationgraphs} learned latent attention-based edges, and recent works such as CASEC \citep{wang2022contextaware} and GACG \citep{duan2024groupawarecoordinationgraphmultiagent} improved scalability and hierarchical reasoning. Mixer-style variants such as GraphMix \citep{naderializadeh2020graph} and QGNN \citep{kortvelesy2022qgnn} combine graph convolution with value mixing. These approaches, however, employ static or per-step relational graphs, limiting adaptability in dynamic settings where coordination structures evolve over time. 
LTSCG~\citep{duan2024inferring} is a closely related work that leverages historical trajectories to infer latent temporal graphs. \algo, however, models temporal dependencies directly within the graph itself, jointly reasoning over structure and time. 


\paragraph{Graph Representation Learning.} GNNs have emerged as a powerful framework for learning over relational data by enabling message passing across structured entities \citep{wu2020comprehensive}. Works such as Graph Attention Networks \citep{veličković2018graphattentionnetworks} and Graph Convolutional Networks \citep{kipf2016semi}, have been widely used to model local dependencies, assign adaptive edge weights, and capture higher-order relationships in dynamic systems. Temporal GNNs jointly model changes in both node attributes and edge structures over time. Notable architectures, such as TGAT \citep{xu2020inductiverepresentationlearningtemporal}, introduce continuous-time encoding and memory modules to efficiently represent long-term dependencies. 
Our work leverages these advances to design a temporal graph learning module that dynamically captures evolving coordination structures to enhance multi-agent policy learning.

\section{Preliminaries } 
\label{sec:preliminaries} 

\paragraph{MARL.} We formalize the problem as a decentralized partially observable Markov Decision Process \citep{oliehoek2016concise}, defined by the tuple $\mathcal{M} = \langle \mathcal{D}, \mathcal{S}, \{\mathcal{A}_i\}_{i \in \mathcal{D}}, \mathcal{P}, \mathcal{R}, \{\mathcal{O}_i\}_{i \in \mathcal{D}}, \Omega, \gamma \rangle,$ where $\mathcal{D}$ denotes the set of $N$ agents. At timestep $t$, the environment is in a global state $s_t \in \mathcal{S}$, and each agent $i \in \mathcal{D}$ selects an action $a_{t,i} \in \mathcal{A}_i$, forming the joint action $\mathbf{a}_t = \langle a_{t,1}, \dots, a_{t,N} \rangle$. The transition function $\mathcal{P}(s_{t+1} | s_t, \mathbf{a}_t)$ defines the probability of reaching $s_{t+1}$, and all agents share a global reward $\mathcal{R}(s_t, \mathbf{a}_t) \in \mathbb{R}$ that quantifies the cooperative objective. Each agent receives a local observation $o_{t,i} \in \mathcal{O}_i$ determined by the joint observation function $\Omega(s_t, \mathbf{a}_t) = \langle o_{t,1}, \dots, o_{t,N} \rangle$. Since agents operate under partial observability, agent $i$ maintains a local action-observation history $\uptau_{t,i} = \{(o_{1,i}, a_{1,i}), \dots, (o_{t,i}, a_{t-1,i})\}$ and acts according to its local policy $\pi_i(a_{t,i} | \uptau_{t,i})$. The joint policy is given by $\boldsymbol{\pi} = \langle \pi_1, \dots, \pi_N \rangle$. The quality of a joint policy $\boldsymbol{\pi}$ is measured by the expected discounted return: $ J(\boldsymbol{\pi}) = \mathbb{E}_{\boldsymbol{\pi}}\!\left[\sum_{t=0}^{\infty} \gamma^t \mathcal{R}(s_t, \mathbf{a}_t)\right],$ where $\gamma \in [0,1)$ is the discount factor. 
The joint action-value function can be defined as: $Q^{\boldsymbol{\pi}}(s_t, \mathbf{a}_t) 
= \mathbb{E}_{\boldsymbol{\pi}}\!\left[\sum_{c=0}^{\infty} \gamma^c \mathcal{R}(s_{t+c}, \mathbf{a}_{t+c}) \mid s_t, \mathbf{a}_t\right].$
The goal in MARL is to find the optimal joint policy $\boldsymbol{\pi}^* = \langle \pi_i^* \rangle_{i \in \mathcal{D}}$ that maximizes this value function: $Q^*(s_t, \mathbf{a}_t) = \max_{\boldsymbol{\pi}} Q^{\boldsymbol{\pi}}(s_t, \mathbf{a}_t).$ Under the centralized training with decentralized execution (CTDE) paradigm \citep{amato2024introduction}, agents exploit global state and joint action information during training to learn coordinated policies, while acting independently using only local trajectories at execution. A common way to realize CTDE is by decomposing the joint value as
\begin{equation}
\label{eq:mixing}
Q_{\mathrm{tot}}
= f_{\psi}\bigl(Q_1(\uptau_{t,1},a_{t,1}),\dots,Q_N(\uptau_{t,N},a_{t,N})\bigr),
\end{equation}
with $f_{\psi}$ a learnable mixing function parameterized by $\psi$. 
Each agent acts greedily with respect to its local utility: 
$a_{t,i} = \arg\max_{a_i} Q_i(\uptau_{t,i}, a_i).$ 


\paragraph{Graph Attention Networks (GAT).} GAT \citep{velivckovic2017graph} provides a flexible way to model relational dependencies through attention-based message passing. In MARL, agents can be represented as nodes and their interactions as edges in a graph $G = (V, E)$, where $V = \{1, \ldots, N\}$ corresponds to the set of agents and $E \subseteq V \times V$ denotes the set of interaction links. Each node $i$ has a feature or embedding vector $h_i \in \mathbb{R}^d$, which aggregates information from its neighbors $\mathcal{N}(i)$. For each neighboring agent $j \in \mathcal{N}(i)$, GAT computes an attention weight $\alpha_{ij}$ that quantifies the importance of agent $j$'s information to agent $i$:
\begin{equation}
\label{eq:attn-weights}
\alpha_{ij} = 
\frac{
    \exp\!\left(\mathrm{LeakyReLU}\!\big(a^{\top} [Wh_i \Vert Wh_j]\big)\right)
}{
    \sum_{k \in \mathcal{N}(i)} 
    \exp\!\left(\mathrm{LeakyReLU}\!\big(a^{\top} [Wh_i \Vert Wh_k]\big)\right)
},
\end{equation}
where $W \in \mathbb{R}^{d \times d'}$ and $a \in \mathbb{R}^{2d'}$ are learnable parameters. The updated embedding for agent $i$ is then computed as $h'_i = \sum_{j \in \mathcal{N}(i)} \alpha_{ij}Wh_j$. 

\paragraph{Temporal Graph Attention Networks (TGAT).} TGAT~\citep{xu2020inductiverepresentationlearningtemporal} extends to evolving graphs by introducing temporal attention, allowing attention weights to depend jointly on structural and temporal context. A temporal graph is defined as \(G = \langle V, E, T \rangle\), where nodes \(V\) represent agents, edges \(E = \{(u,v,t)\}\) are time-stamped interactions, and \(T\) is the continuous time domain. Each node \(v_i\) maintains a feature vector \(x_i \in \mathbb{R}^{d_0}\) and a time-dependent embedding \(h_i(t)\). TGAT constructs an entity-temporal feature matrix for each node \(v_i\) at time \(t\):
\begin{equation}
Z_i(t) =
[\,h_i(t)\Vert \phi(0);\,
(h_j(t_j)\Vert \phi(t-t_j))_{v_j \in \mathcal{N}(v_i,t)}]^{\!\top},
\end{equation}
where \(\phi(\cdot)\in\mathbb{R}^{d_T}\) encodes relative timestamps. Linear projections of \(Z_i(t)\) yield the query, key, and value matrices:
\(
q_i = Z_{i,0}W_q,\;
K_i = Z_{i,1:}W_K,\;
V_i = Z_{i,1:}W_V,
\)
with \(W_q, W_K, W_V \in \mathbb{R}^{(d_0+d_T)\times d_h}\). Temporal attention over neighbors is then computed as
\begin{equation}
\alpha_{ij}(t) =
\frac{\exp(q_i^\top K_{ij})}
{\sum_{v_k\in\mathcal{N}(v_i,t)} \exp(q_i^\top K_{ik})},
\end{equation}
Temporally weighted aggregation produces the updated embedding: $h_i(t) = \sum_{v_j\in\mathcal{N}(v_i,t)} \alpha_{ij}(t)\,V_{ij}.$ This formulation allows TGAT to selectively emphasize temporally and structurally relevant interactions, yielding time-aware node embeddings well-suited for dynamic relational reasoning.

\section{Methodology} 
\label{sec:methodology} 


This section presents the architecture of \algo\ for MARL. We first construct dynamic temporal graphs to capture both static and temporal inter-agent relationships. We then apply temporal attention to aggregate information from these graphs, obtaining agent embeddings that reflect evolving coordination. Finally, we integrate these with MARL to improve their performance. 


\paragraph{Dynamic Temporal Graph Construction.} To capture both current and evolving coordination structures among agents, \algo\ constructs a temporal graph at each timestep that combines three distinct types of connections: 
(i) \textit{static neighbors} representing current interactions, 
(ii) \textit{self-history} edges that link an agent to its own past states, and 
(iii) \textit{neighbor-history} edges connecting it to the past states of its current neighbors. 
\begin{figure*}[!t]
  \centering
  \includegraphics[width=0.75\linewidth]{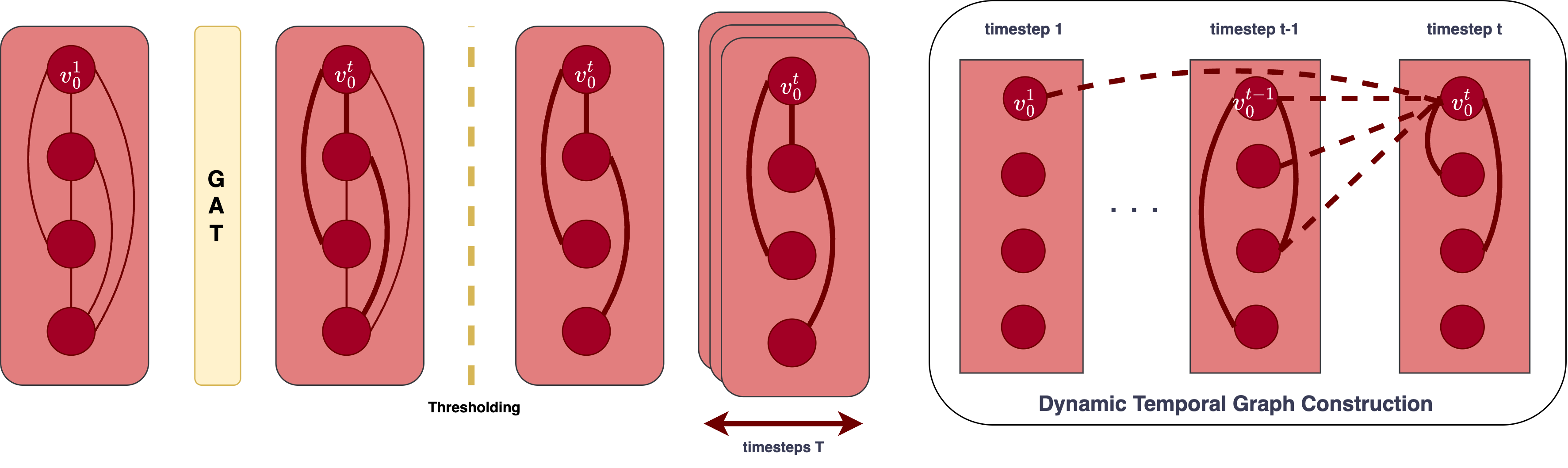}
  \caption{\textbf{Dynamic temporal graph construction in \algo.} At each timestep $t$, GAT-pruned structural edges (\textit{static neighbors}) are augmented with \textit{self-} and \textit{neighbor-history} links from the past $K_{\mathrm{past\_self}}$ and $K_{\mathrm{past\_nbr}}$ steps, capturing both current and evolving coordination among agents.} 
  \label{fig:tg_constr}
\end{figure*}
As illustrated in Fig.~\ref{fig:tg_constr}, for each timestep $t$, we begin with a fully connected graph $G_t = \langle V_t, E_t \rangle$ over the $N$ agents, where $V_t = \{v_1^t, \ldots, v_N^t\}$ and $|E_t| = N(N-1)/2$. A GAT layer assigns attention weights $\alpha_{ij}^t$ to all edges, indicating the relative importance of agent $j$ to agent $i$ at time $t$. To retain only the most significant interactions, we prune the graph by selecting the top $K_{\mathrm{stat\_nbr}}$ fraction of edges with the highest attention scores, forming a sparse set of structural connections $E_t^{\mathrm{static}} = K_{\mathrm{stat\_nbr}} \times |E_t|$. The resulting static neighborhood of agent $i$ at time $t$ is thus: $\mathcal{N}_t(v_i^t) = \{\,v_j^t \;|\; e_{ij}^t \in E_t^{\mathrm{static}}\,\}.$ Next, we incorporate temporal information by augmenting this static graph with historical connections that capture how coordination evolves across time. Each agent is linked to its own past instances up to a fixed horizon $K_{\mathrm{past\_self}}$: $\mathcal{N}^{\mathrm{self}}(v_i^t) = \{\,v_i^{t-\Delta} \mid 1 \leq \Delta \leq K_{\mathrm{past\_self}}\,\},$ and to the past states of its static neighbors up to $K_{\mathrm{past\_nbr}}$ timesteps: $\mathcal{N}^{\mathrm{nbr}}(v_i^t) = \{\,v_j^{t-\Delta} \mid v_j^{t} \in \mathcal{N}_t(v_i^{t}),\, 1 \leq \Delta \leq K_{\mathrm{past\_nbr}}\,\}.$ The complete temporal neighborhood of agent $i$ at time $t$ thus becomes: $\mathcal{N}(v_i^t) = \mathcal{N}_t(v_i^t) \cup \mathcal{N}^{\mathrm{self}}(v_i^t) \cup \mathcal{N}^{\mathrm{nbr}}(v_i^t).$ This formulation provides explicit control over how much historical information is integrated into the temporal graph through the parameters $K_{\mathrm{stat\_nbr}}$, $K_{\mathrm{past\_self}}$, and $K_{\mathrm{past\_nbr}}$. Formally, we can define a \textit{static-temporal trade-off} that constrains the size of an agent's temporal neighborhood as:
\begin{equation}
|\mathcal{N}(v_i^t)|
  = K_{\mathrm{past\_self}}
  + \big(K_{\mathrm{stat\_nbr}} \times |E_t|\big)
  + \big(K_{\mathrm{stat\_nbr}} \times |E_t|\big) \times K_{\mathrm{past\_nbr}}.
\label{eq:nbr_size}
\end{equation}
This governs how the temporal and structural components interact, determining how much of the agents’ history is leveraged relative to current connectivity. In Section~\ref{sec:results}, we show resulting temporal graph enhances coordination performance. Through targeted ablations, we further examine how varying each component shapes learning dynamics. Analysis of computational complexity and graph size is provided in Appendix~\ref{app:complexity}.

\begin{figure*}[!t]
  \centering
  \includegraphics[width=0.75\textwidth]{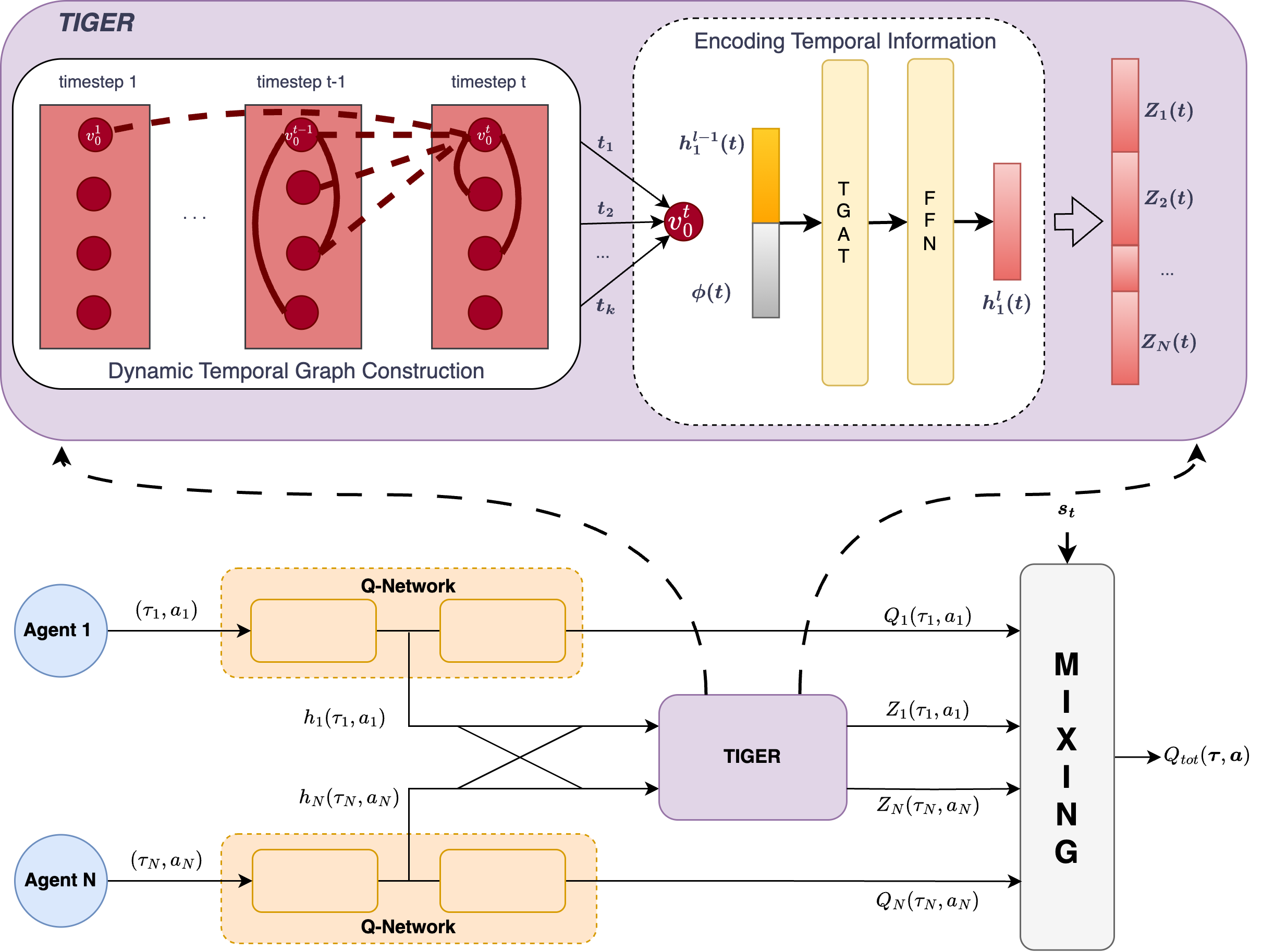} 
  \caption{\textbf{Overview of \algo.} At each timestep, \algo\ constructs a dynamic temporal graph capturing both current and evolving inter-agent dependencies. Temporal attention then encodes these graphs into rich, weighted agent embeddings that integrate the most informative relational signals. 
  These time-aware agent embeddings guide the learning of agent policies in MARL.} 
  \label{fig:tiger}
\end{figure*}

\paragraph{Temporal Graph Encoding.} Once the dynamic temporal graph is constructed, the next step is to derive time-aware agent feature representations. For this, \algo\ includes a temporal encoder inspired by TGAT~\citep{xu2020inductiverepresentationlearningtemporal}, given the proven effectiveness of attention mechanisms in TGNNs. This enables agents to dynamically incorporate information from recent and past interactions, focusing on the most relevant connections through temporal attention. For each agent $v_i$ at time $t$, we construct an entity-temporal feature matrix (as introduced in Section~\ref{sec:preliminaries}) that concatenates its current embedding with the temporally encoded features of its neighbors: $Z_i(t) = \bigl[\, h_i^{(0)}(t)\Vert\phi(0);\, (h_j^{(0)}(\tau)\Vert\phi(t-\tau))_{v_j^\tau \in \mathcal{N}(v_i^t)} \bigr]^{\!\top},$ where $\phi:\mathbb{R}_{\ge 0}\!\to\!\mathbb{R}^{d_T}$ is a continuous-time encoder that maps relative timestamps to temporal embeddings. The rows of $Z_i(t)$ are linearly projected into a shared latent space using $W_q, W_K, W_V \!\in\! \mathbb{R}^{(d_0+d_T)\times d_h}$, producing the query, key, and value matrices: $ q_i = Z_{i,0} W_q,\quad K_i = Z_{i,1:} W_K,\quad V_i = Z_{i,1:} W_V.$ Temporal attention weights are then computed as $\alpha_{j\tau} = \frac{\exp(q_i^{\top} K_{j\tau})}{\sum_{v_k^{\nu}\in\mathcal{N}(v_i^t)} \exp(q_i^{\top} K_{k\nu})},$ assigning higher importance to temporally and structurally relevant neighbors. The agent aggregates the contextual information from its temporal neighborhood as $\tilde{h}_i(t) = \sum_{v_j^{\tau} \in \mathcal{N}(v_i^t)} \alpha_{j\tau} V_{j\tau}.$ Finally, it fuses this temporal message with its current observation through a feed-forward transformation: $h_i(t) = \mathrm{ReLU}\!\bigl([\tilde{h}_i(t)\Vert o_{t,i}]W_0+b_0\bigr)W_1+b_1,$ where $W_0$, $W_1$, $b_0$, and $b_1$ are learnable parameters. The resulting embeddings $h_i(t)$ encode both structural and temporal dependencies, enriching each agent’s feature representation.

\paragraph{Integration with MARL.} The updated embeddings $h_i(t)$ produced by the encoder serve as input features for each agent’s value estimation in CTDE (see Section~\ref{sec:preliminaries}). By integrating information from the constructed temporal graph, these embeddings provide a latent representation that captures the evolving coordination context, enabling agents to make more informed decisions. Each agent $i$ maintains a local utility head that estimates the value of taking action $a_{t,i}$ given its temporal embedding and the global state: \(\;Q_i\bigl(s_t \Vert h_i(t),a_{t,i}\bigr)\). These local Q-values are combined through a mixing network $f_{\psi}$ to form the joint action-value function,
\begin{equation}
Q_{\mathrm{tot}}(s_t, \mathbf{a}_t)
  = f_{\psi}\!\big(Q_1, \ldots, Q_N;\, s_t \Vert [h_1(t), \ldots, h_N(t)]\big),
\label{eq:qtot_tiger}
\end{equation}
which follows the standard value decomposition form of Eq.~\ref{eq:mixing}. 
This formulation allows temporal reasoning to enhance centralized value estimation during training without affecting decentralized execution at test time. During deployment, each agent selects its action independently using its updated embedding: $a_{t,i} = \arg\max_{a_i}\, Q_i(a_i \mid \uptau_{t,i}, h_i(t)).$ 
We demonstrate that this design improves task performance and sample efficiency in cooperative MARL tasks (Section~\ref{sec:results}).

\section{Experimental Setup} 
\label{sec:experiments}

\paragraph{Environments.} We evaluate \algo\ on two popular cooperative multi-agent tasks, Gather \citep{wang2022contextaware} and Tag \citep{lowe2017multi}. These are coordination-intensive and test complementary aspects of temporal reasoning and adaptability in dynamic environments (see Appendix~\ref{app:tasks}). 

\begin{itemize}[nolistsep,leftmargin=*]
    \item \textit{Gather} extends Climb \citep{wei2016lenient} to induce delayed cooperation under partial observability. Each agent selects among three actions $\{a_0,a_1,a_2\}$ for goals $\{g_1,g_2,g_3\}$, with one goal randomly chosen as optimal and known only to nearby agents. A team reward of $+10$ is given if all reach the optimal goal, $+5$ for a non-optimal one, and $-5$ if only a subset succeeds. This setup requires agents to infer and share latent information over time to coordinate effectively. 
    \item \textit{Tag} is a continuous-space particle-world task based on the Multi-Agent Particle Environment \citep{lowe2017multi}, where a team of slower pursuers must collaboratively capture faster evaders in the presence of obstacles. Agents receive shared rewards upon successful captures, making cooperation essential for success. With constantly shifting spatial and relational agent dependencies, Tag challenges the need for real-time adaptation of coordination strategies. 
\end{itemize}




\paragraph{Baselines.} We compare \algo\ against two representative algorithm families in cooperative MARL: 


\begin{itemize}[nolistsep,leftmargin=*]
    \item \textit{Mixers:} This family captures approaches that aggregate per-agent utilities through a joint value function without explicitly modeling coordination structure. We include VDN \citep{sunehag2018value} as a minimal factorization baseline and QMIX \citep{rashid2018qmix} as its monotonic generalization. We also evaluate against GraphMix \citep{naderializadeh2020graph} and QGNN \citep{kortvelesy2022qgnn}, which incorporate message passing among agents prior to value aggregation. Together, these baselines test whether \algo\ complements or subsumes relational information implicitly captured by mixer-style architectures. 
    \item \textit{Graph-based methods:} This includes algorithms that construct and learn coordination graphs to represent inter-agent dependencies. We consider DICG \citep{li2021deepimplicitcoordinationgraphs}, which learns attention-weighted graphs; CASEC \citep{wang2022contextaware}, which introduces sparsity through variance-based edge pruning; GACG \citep{duan2024groupawarecoordinationgraphmultiagent}, which dynamically groups agents to capture higher-order relations; and LTSCG \citep{duan2024inferring}, which infers temporal sparse graphs from historical observations. This diverse set enables us to isolate how explicit temporal graph updates in \algo\ enhance MARL under evolving interaction patterns. More details are in Appendix~\ref{app:implementation}. 
\end{itemize}


\section{Experiments and Results} 
\label{sec:results} 

This section evaluates our approach on cooperative MARL tasks introduced in Section~\ref{sec:experiments}. Section~\ref{sec:mainresults} presents the main findings, highlighting the benefits of temporal relational reasoning through comparisons with strong mixer- and graph-based baselines. Section~\ref{sec:ablations} then analyzes how specific design choices in \algo\ influence these results through targeted ablations.


\subsection{Main results} 
\label{sec:mainresults} 


In this section, we demonstrate that incorporating temporal relational reasoning in MARL improves performance by comparing \algo\ against several MARL baselines. We implement \algo\ as a modular component that augments diverse MARL learners. Specifically, we integrate it into two baselines (QMIX and DICG) representing the two families identified in Section~\ref{sec:experiments}. This yields two \algo\ variants: \algo-MIX and \algo-DICG. All methods are evaluated through two common metrics: (i) \textit{task performance}, measured as test win rate in Gather and test return in Tag; and (ii) \textit{sample efficiency}, defined as the number of steps required to reach convergence. Each result is averaged over five independent trials, with the mean and standard deviation reported. We instantiate $K_{\mathrm{past\_nbr}}=1$, $K_{\mathrm{past\_self}}=1$, and $K_{\mathrm{static\_nbr}}=50\%$ in Eq. \ref{eq:nbr_size}, preserving a compact temporal neighborhood of size $N+1$. Further effects of varying these parameters are analyzed in Section~\ref{sec:ablations}.  

\begin{figure*}[!t]
  \centering
  \setcounter{subfig}{0}

  \begin{minipage}[b]{0.49\textwidth}
    \centering
    \refstepcounter{subfig}\label{fig:main-results-a} 
    \includegraphics[width=\linewidth]{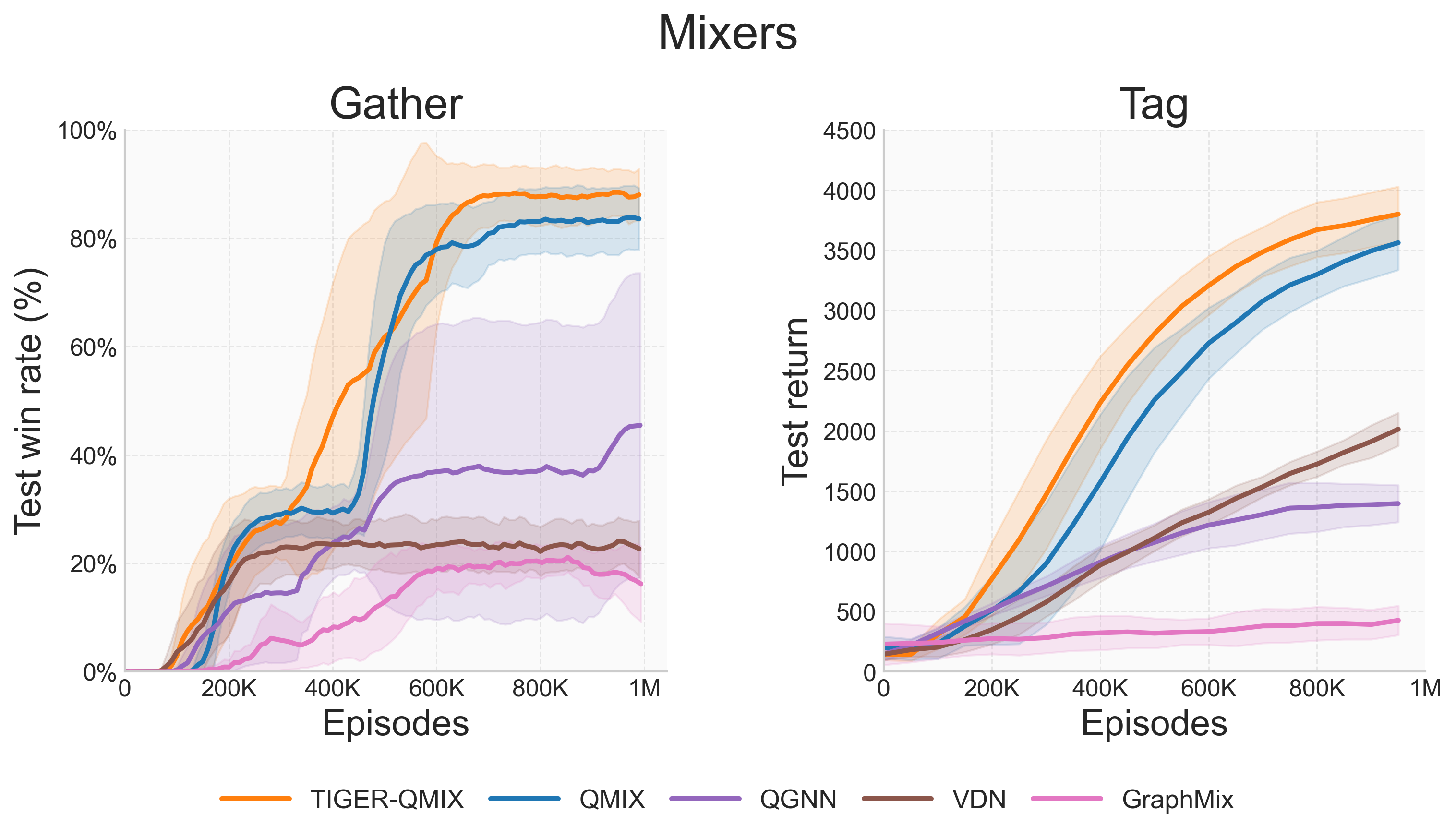}
    \\[-1ex](\alph{subfig})
  \end{minipage}
  \hfill
  \begin{minipage}[b]{0.49\textwidth}
    \centering
    \refstepcounter{subfig}\label{fig:main-results-b} 
    \includegraphics[width=\linewidth]{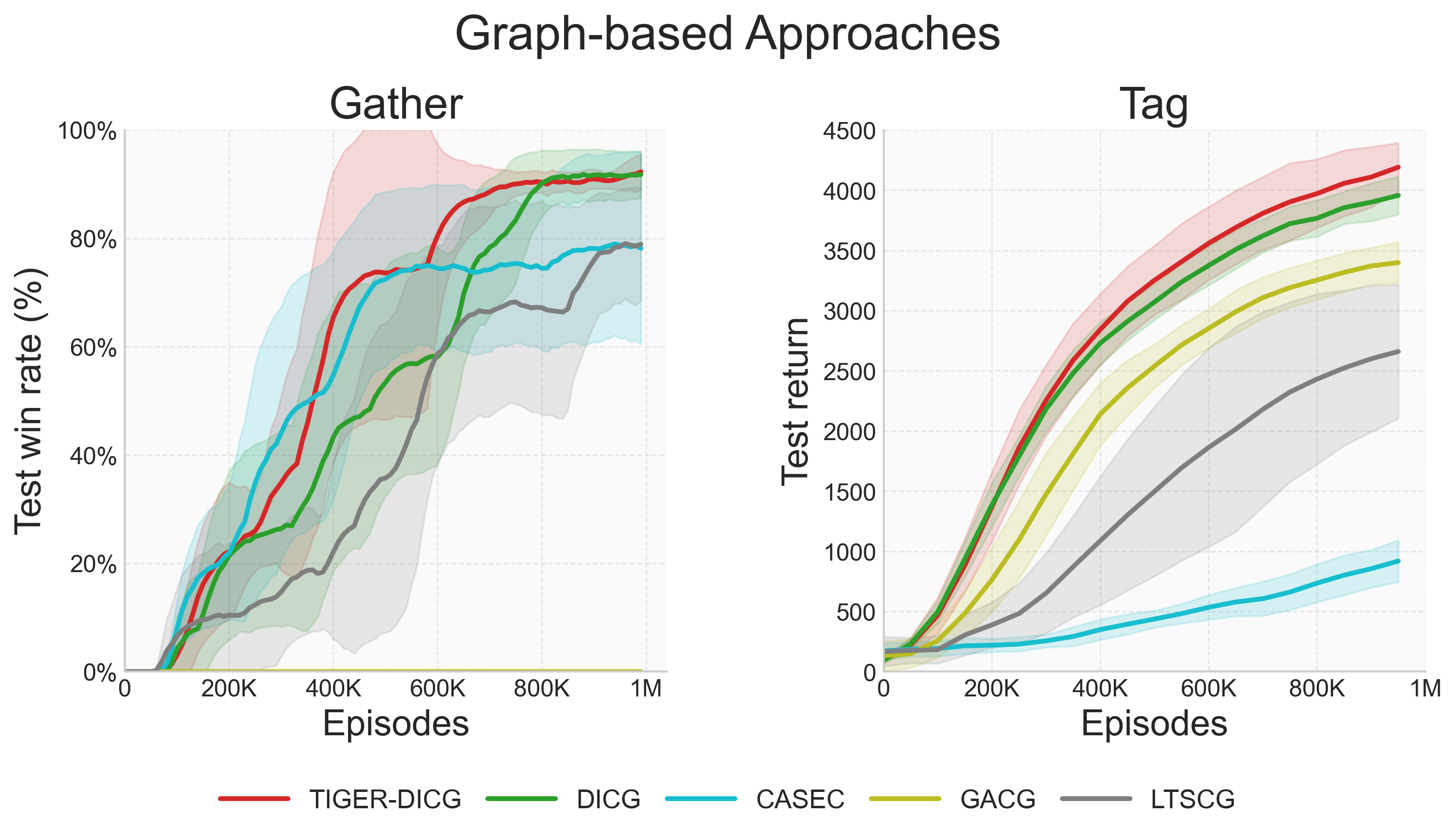}
    \\[-1ex](\alph{subfig})
  \end{minipage}

  \caption{\textbf{Performance of \algo on Gather and Tag.} 
  Plots show task performance (win rate for Gather, return for Tag) over training steps, with standard deviation as shaded regions.  
  (a) Mixer-based results showing faster convergence and stronger asymptotic performance with \algo-MIX.
  (b) Graph-based results showing more stable learning and higher final returns with \algo-DICG.}
  \label{fig:main-results}
\end{figure*}

\paragraph{Comparison with Mixers.} 
We begin by evaluating \algo-MIX, where \algo\ augments QMIX's mixing network. Each agent's hidden state is embedded within \algo's temporal graph, allowing information to propagate across recent interactions and yielding temporally enriched features for value decomposition. Across both Gather and Tag, \algo-MIX consistently outperforms mixer baselines (Fig.~\ref{fig:main-results-a}). The gains in performance and sample efficiency are most pronounced in Gather. Here, \algo-MIX converges faster to a higher asymptotic win rate than all baselines, indicating the benefits of capturing evolving agent information. In Tag, \algo-MIX maintains a clear margin in return and exhibits lower variance across runs. These results are particularly meaningful, as QMIX enforces a monotonic value factorization, GraphMix integrates attention-based graph mixing to model static dependencies, and QGNN employs multi-layer message passing to capture spatial relationships. 
Outperforming these baselines suggests that \algo’s temporal graph modeling captures dependencies that neither purely value-based nor static-graph mixers can express. 

\paragraph{Comparison with Graph-based methods.} 
We next evaluate \algo-DICG, which extends DICG by enriching its coordination graphs with \algo's temporal information modeling. Compared to DICG, CASEC, GACG, and LTSCG (Fig.~\ref{fig:main-results-b}), \algo-DICG consistently achieves higher task performance and better sample efficiency across both environments. On Gather, the improvement in win rate is substantial compared to CASEC, GACG, and LTSCG and modest in absolute value compared to DICG, but it occurs considerably earlier. The gains are more pronounced in \textit{Tag}. \algo-DICG converges faster to higher final returns. These results demonstrate that temporal reasoning complements graph-based MARL methods in cooperative settings. CASEC builds sparse coordination topologies based on payoff variance but lacks temporal continuity. GACG captures group-level dependencies while remaining static over time, and LTSCG constructs latent temporal graphs from historical trajectories, making its temporal reasoning indirect through inferred structures. Whereas \algo-DICG directly integrates temporal edges into an evolving coordination graph.

\subsection{Ablations}
\label{sec:ablations}

\begin{figure}[!t]
  \centering
  \setcounter{subfig}{0}

  \begin{minipage}{.33\textwidth}
    \centering
    \refstepcounter{subfig}\label{fig:study1}
    \includegraphics[width=\linewidth]{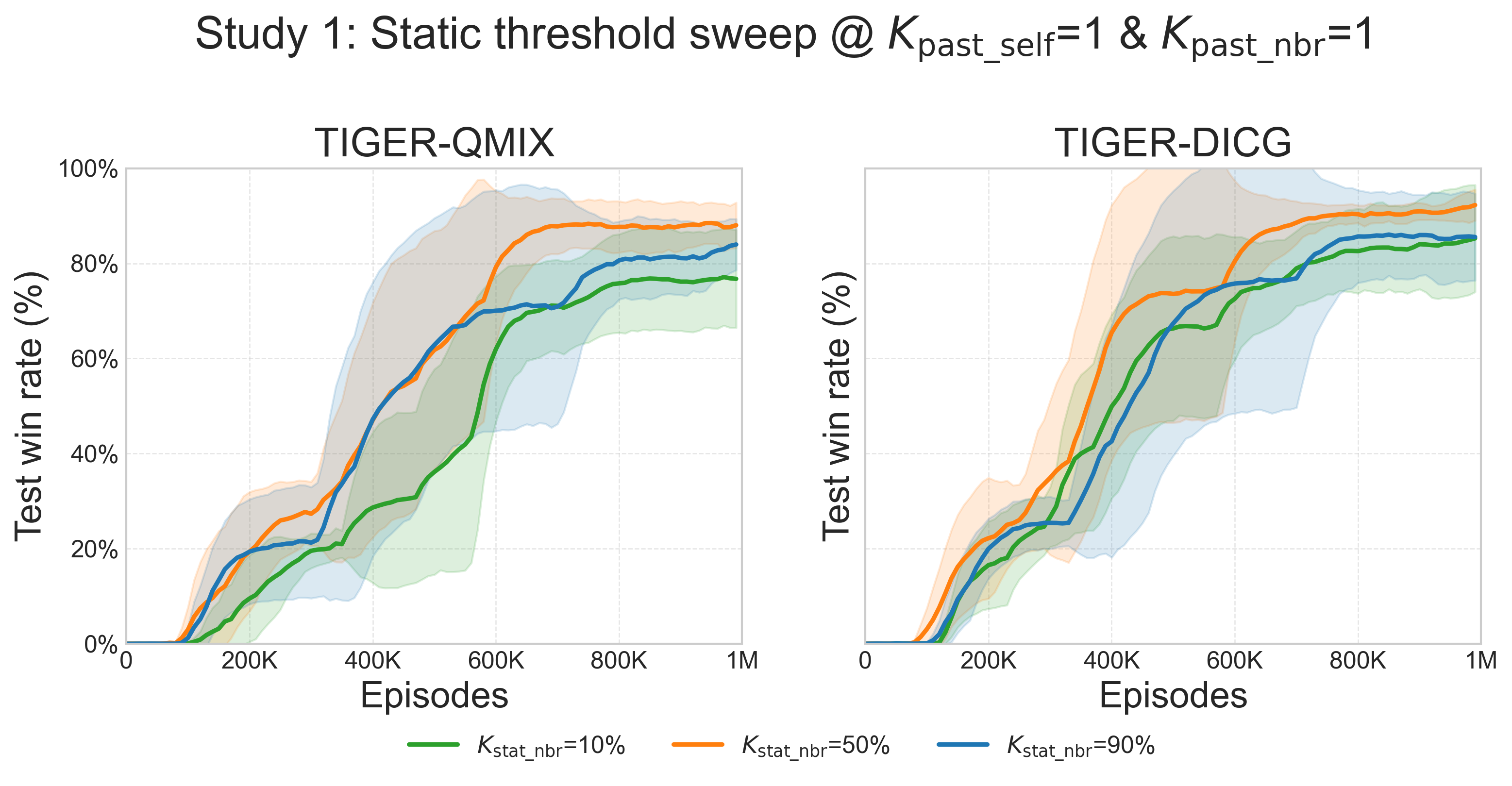}
    \\[-1ex](\alph{subfig})
  \end{minipage}%
  \hfill
  \begin{minipage}{.33\textwidth}
    \centering
    \refstepcounter{subfig}\label{fig:study2}
    \includegraphics[width=\linewidth]{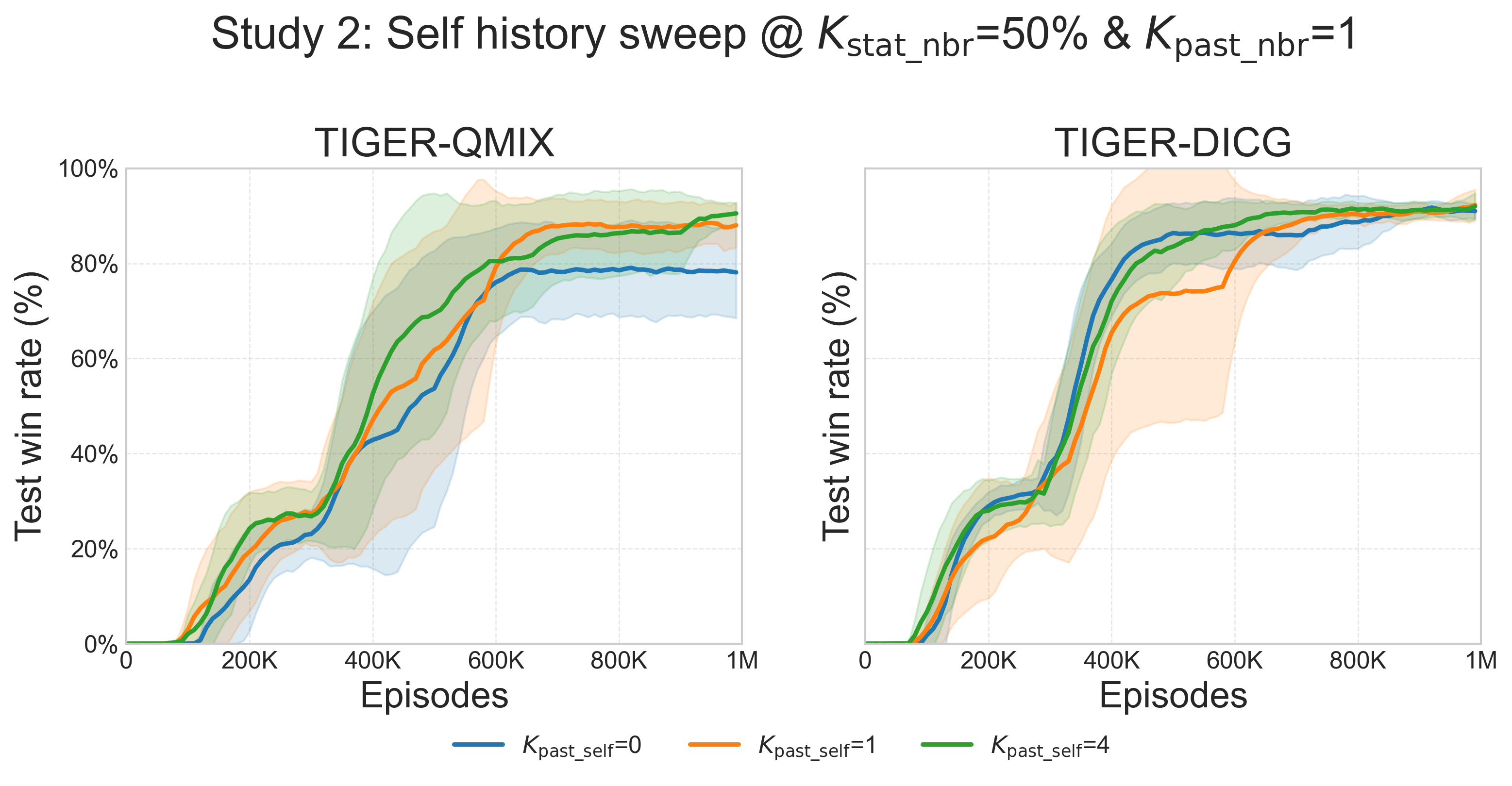}
    \\[-1ex](\alph{subfig})
  \end{minipage}%
  \hfill
  \begin{minipage}{.33\textwidth}
    \centering
    \refstepcounter{subfig}\label{fig:study3}
    \includegraphics[width=\linewidth]{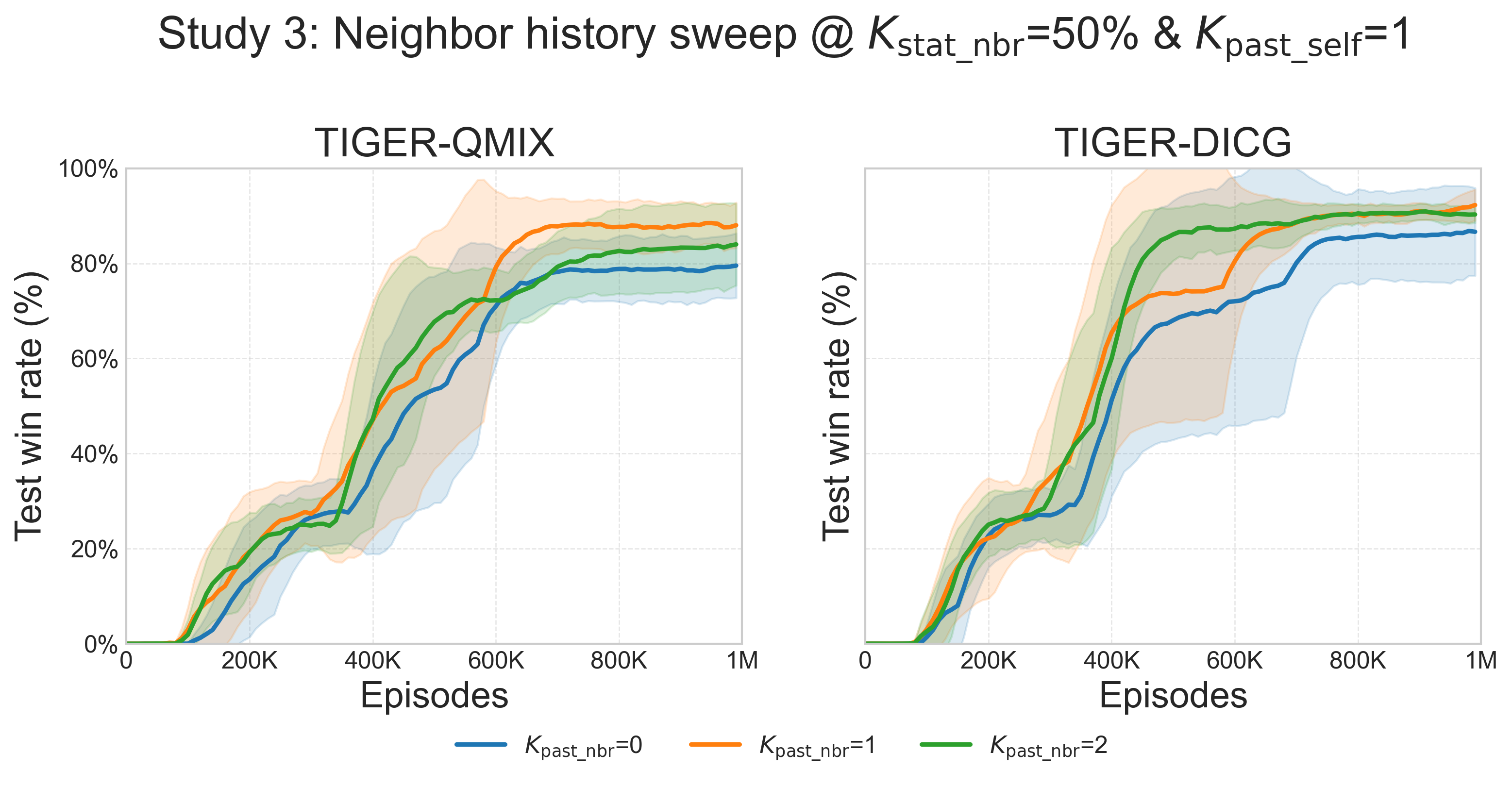}
    \\[-1ex](\alph{subfig})
  \end{minipage}

\caption{\textbf{Ablation studies on \algo.} 
We vary (a) static connectivity ($K_{\mathrm{stat\_nbr}}$), (b) self-history depth ($K_{\mathrm{past\_self}}$), and (c) neighbor-history depth ($K_{\mathrm{past\_nbr}}$) to analyze how structural and temporal factors affect \algo’s performance.
}
\label{fig:ablations}
\end{figure}

In this section, we study how key design choices in \algo\ ($K_{\mathrm{stat\_nbr}}$, $K_{\mathrm{past\_self}}$, and $K_{\mathrm{past\_nbr}}$ in Eq.~\ref{eq:nbr_size}) affect its overall performance. 
While the main results validate \algo\ across benchmarks, this section focuses on an in-depth analysis using Gather, which offers stable convergence and clear performance separability. 
Tag shows subtle but qualitatively similar trends (see Appendix~\ref{app:tag-ablations}).


\paragraph{Study 1: Static neighbors $K_{\textrm{static\_nbr}}$ (Fig.~\ref{fig:study1}).} Starting from the configuration introduced in Section~\ref{sec:mainresults} ($K_{\textrm{static\_nbr}}=50\%$), we first reduce to a very sparse setting ($K_{\textrm{static\_nbr}}=10\%$). Performance drops noticeably, indicating that without sufficient structural edges, \algo's temporal graph has limited context to propagate information. This shows that \algo’s benefits rely on the added graph-based agent information integration rather than added network capacity in the module. Next, as we approach nearly complete graph ($K_{\textrm{static\_nbr}}=90\%$), performance does not improve further or even drops slightly. We speculate that excessive or redundant edges introduce noisy or conflicting signals, and since $K_{\textrm{static\_nbr}}$ also influences temporal edge formation, oversaturation amplifies such effects. Overall, a moderate level of static connectivity gives the most effective performance. 

\paragraph{Study 2: Self history $K_{\textrm{past\_self}}$ (Fig.~\ref{fig:study2}).} To examine the influence of temporal recall over an agent's own past observations, we remove self-history edges ($K_{\textrm{past\_self}}=0$). The performance of \algo-MIX drops, showing that short-term self-history is vital, while \algo-DICG remains relatively stable. We then increase the setting to a higher temporal depth. To maintain computational efficiency, we follow a logarithmic rule $K_{\textrm{past\_self}}=\lceil\ln(NT)\rceil$. This scales gracefully across environments, and ensures sublinear growth in self-history edges ($\mathcal{O}(NT)$). It yields values of $4$ for Gather and $7$ for Tag (see Appendix~\ref{app:complexity}). In Gather, $K_{\textrm{past\_self}}=4$ offers only marginal benefits. This suggests that short-term past states already encode most relevant information, or other parameters (current or past neighbors) dominate or compensate for the contribution of deeper self-history. 

\paragraph{Study 3: Neighbor history $K_{\textrm{past\_nbr}}$ (Fig.~\ref{fig:study3}).} Upon removing past-neighbor links ($K_{\textrm{past\_nbr}} = 0$), the performance drops noticeably across both \algo\ variants. This confirms that recalling recent neighbor interactions is crucial in \algo. When extended to two-hop temporal neighbors, the improvement becomes marginal or even slightly drops, suggesting that most of the useful relational information is already captured by a single temporal hop. Thus, further increases are likely to provide limited new information. Also, increasing temporal neighbor depth introduces additional computational cost, scaling as $\mathcal{O}(N^2T)$, which makes larger hops impractical for many applications. Overall, short-range neighbor recall (one to two hops) provides sufficient temporal context for effective coordination while keeping computation manageable.

\begin{figure}[!t]
  \centering
  \setcounter{subfig}{0}

  \begin{minipage}{0.49\textwidth}
    \centering
    \refstepcounter{subfig}\label{fig:study4-qmix}
    \includegraphics[width=\linewidth]{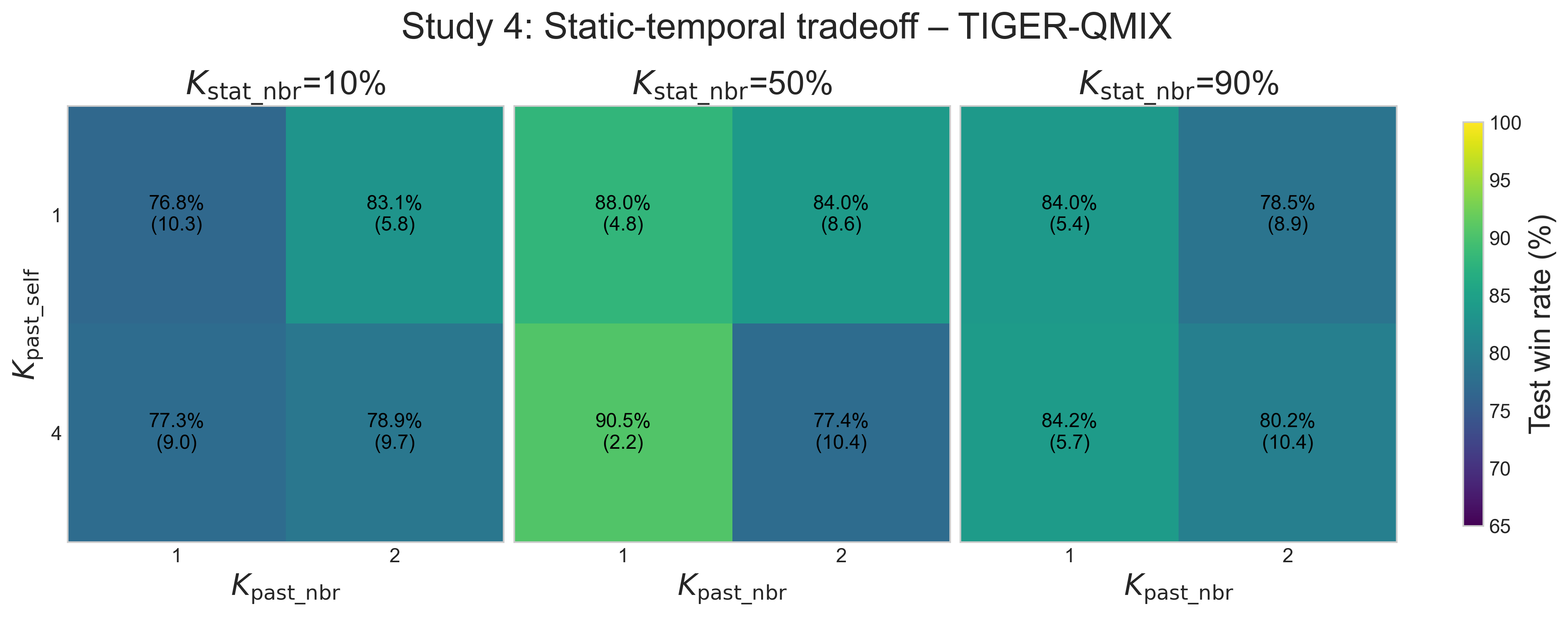}
    \\[-1ex](\alph{subfig})
  \end{minipage}%
  \hfill
  \begin{minipage}{0.49\textwidth}
    \centering
    \refstepcounter{subfig}\label{fig:study4-dicg}
    \includegraphics[width=\linewidth]{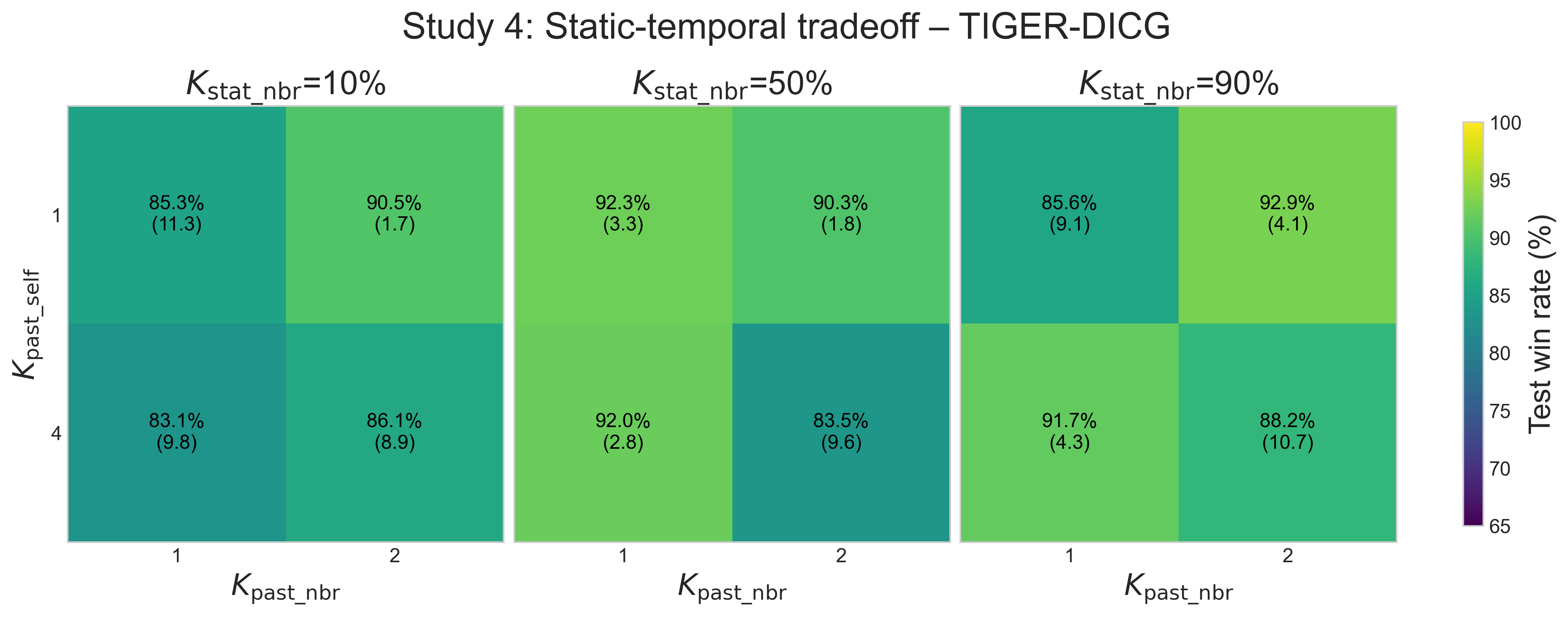}
    \\[-1ex](\alph{subfig})
  \end{minipage}
\caption{\textbf{Static-temporal tradeoff in \algo.} Plots report the final win rate at 1M training steps and standard deviation across runs in parentheses. Each configuration jointly varies static and temporal parameters to understand how they interact in shaping coordination. 
\textbf{(a)} \algo-MIX peaks at moderate static connectivity (50\%) and shallow temporal depth (one-hop)
\textbf{(b)} \algo-DICG remains robust across most settings, with 50\% static threshold offering the best balance.}
  \label{fig:study4:tradeoff}
\end{figure}

\paragraph{Study 4: Static-temporal tradeoff (Fig.~\ref{fig:study4:tradeoff}).} Finally, we jointly vary static and temporal parameters to understand how they interact in shaping coordination. Overall, although win rates fluctuate within a narrow band ($\approx$76-91\%), we observe static and temporal components complementing each other. When structural connectivity is sparse, adding temporal links improves stability (low standard deviation) and performance (high win rate), effectively recovering missing relational cues. However, in dense graphs, deeper temporal recall provides little or even negative benefit, likely due to redundant signals blurring coordination. Moderate connectivity ($50\%$) with shallow temporal depth (one-hop) consistently provides the best balance. 
Notably, \algo-DICG remains stable across configurations, while \algo-QMIX shows higher sensitivity; expected as QMIX does not model any interactions without \algo\ unlike DICG. \\


\noindent In this section, we examined how key design parameters in \algo\ influence its performance. The results show that \algo’s improvements arise primarily from temporal reasoning rather than added capacity (Studies 2, 3), with self and neighbor histories providing informative temporal context. At the same time, the interaction between adjustable parameters highlights the complementary role of static and temporal components in shaping coordination dynamics (Studies 1, 4).

\section{Conclusions} 
\label{sec:conclusion} 

This work introduced \algo, a temporal graph learning framework that enhances cooperative MARL by explicitly modeling how inter-agent relationships evolve over time. By constructing dynamic temporal graphs and encoding them through temporal attention, \algo\ produces time-aware agent embeddings that integrate both structural and temporal context. Our experiments demonstrate that this temporal reasoning substantially improves coordination efficiency, convergence speed, and overall task performance across diverse MARL paradigms, including value-decomposition and graph-based methods. Extensive ablations further reveal that lightweight temporal recall and balanced structural connectivity are sufficient for stable, high-performing coordination. 
Future work could explore deploying \algo\ in various domains such as intelligent traffic management, connected autonomous vehicles, and distributed robotic teams, where agents must coordinate under dynamic and partially observable conditions.  


\bibliography{refs}

\newpage 
\section*{Appendix} 
\appendix 
\label{sec:appendix} 





\noindent\rule{\linewidth}{1pt}

\noindent \textbf{Appendix A:} Additional details on task settings. 

\noindent \textbf{Appendix B:} Further experiments and ablations on Tag. 

\noindent \textbf{Appendix C:} Implementation details. 

\noindent \textbf{Appendix D:} Discussion on computational complexity. 

\noindent\rule{\linewidth}{1pt} 

\section{Additional details on the task settings} 
\label{app:tasks}

We elaborate on high-level environment overview from Section~\ref{sec:experiments}.

\paragraph{Gather \citep{wang2022contextaware}.} Gather is an extension of the classic Climb Game \citep{wei2016lenient}, where each agent chooses among three actions \( A = \{a_0, a_1, a_2\} \). Action \(a_0\) yields no immediate reward unless selected simultaneously by all agents, in which case it results in a high collective reward of \(+10\). Actions \(a_1\) and \(a_2\) provide smaller individual rewards \((+5)\) independently. The task requires strong coordination among agents to consistently select the cooperative action. We adopt the precise setup defined by the Multi-Agent Coordination (MACO) benchmark \citep{wang2022contextaware} to facilitate comparison. Episodes last at most eight steps. Gather features a particularly sparse and delayed reward structure: agents must both identify a hidden target and synchronously occupy it before any payoff is received. This sparsity makes credit assignment challenging, as individual exploratory actions yield no immediate feedback, forcing agents to infer and align their teammates’ intentions over multiple time steps in order to secure the collective \(+10\) bonus.

\paragraph{Tag \citep{lowe2017multi}.} Tag is an environment based on the particle world scenario, in which a team of agents collaboratively chases multiple adversaries within a map containing randomly generated obstacles. The motion of each agent is discretized into five macroactions: \emph{no‑op}, move \(\pm x\), or move \(\pm y\). Episodes last up to 100 environment steps. Agents earn a global reward each time they successfully collide with an adversary. Specifically, every time an agent collides with the adversary, \emph{all} agents instantly receive \(+1\); simultaneous tags sum, and there are no penalties for failed attempts or wall impacts. Furthermore, adversaries have a speed advantage, and so successful captures typically require pursuers to assume complementary roles (for instance, blocking versus chasing), making the task a test of real-time coordination under dense yet still globally shared feedback. We used the experimental setup with ten agents pursuing three adversaries. In Tag, rewards are far more frequent but arise in a dynamic, partially observable setting: agents must implicitly divide roles - such as blocking escape routes versus direct pursuit - according to the evader’s maneuvers and the obstacle layout. The dense feedback accelerates learning, yet optimal performance still hinges on adaptive cooperation and rapid role reassignment in response to adversary behavior. \\

\noindent Gather features a sparse and delayed reward structure, requiring agents to jointly identify and occupy a hidden target. This setup challenges coordination, as agents must infer each other's intentions without immediate feedback. Tag, however, provides dense rewards in a dynamic, partially observable environment, but still demands real-time cooperation as agents must implicitly assume complementary roles (e.g., blocking vs. chasing) based on adversary behavior and obstacles. These environments reflect distinct coordination demands; consistent gains across both demonstrate \algo's effectiveness under varied credit assignment challenges (see Section~\ref{sec:results}).

\section{Further Experiments and Ablations on Tag} 
\label{app:tag-ablations}

\begin{figure}[!t]
  \centering
  \setcounter{subfig}{0}
  \begin{minipage}{.32\textwidth}
    \centering
    \refstepcounter{subfig}\label{fig:tag_study1}
    \includegraphics[width=\linewidth]{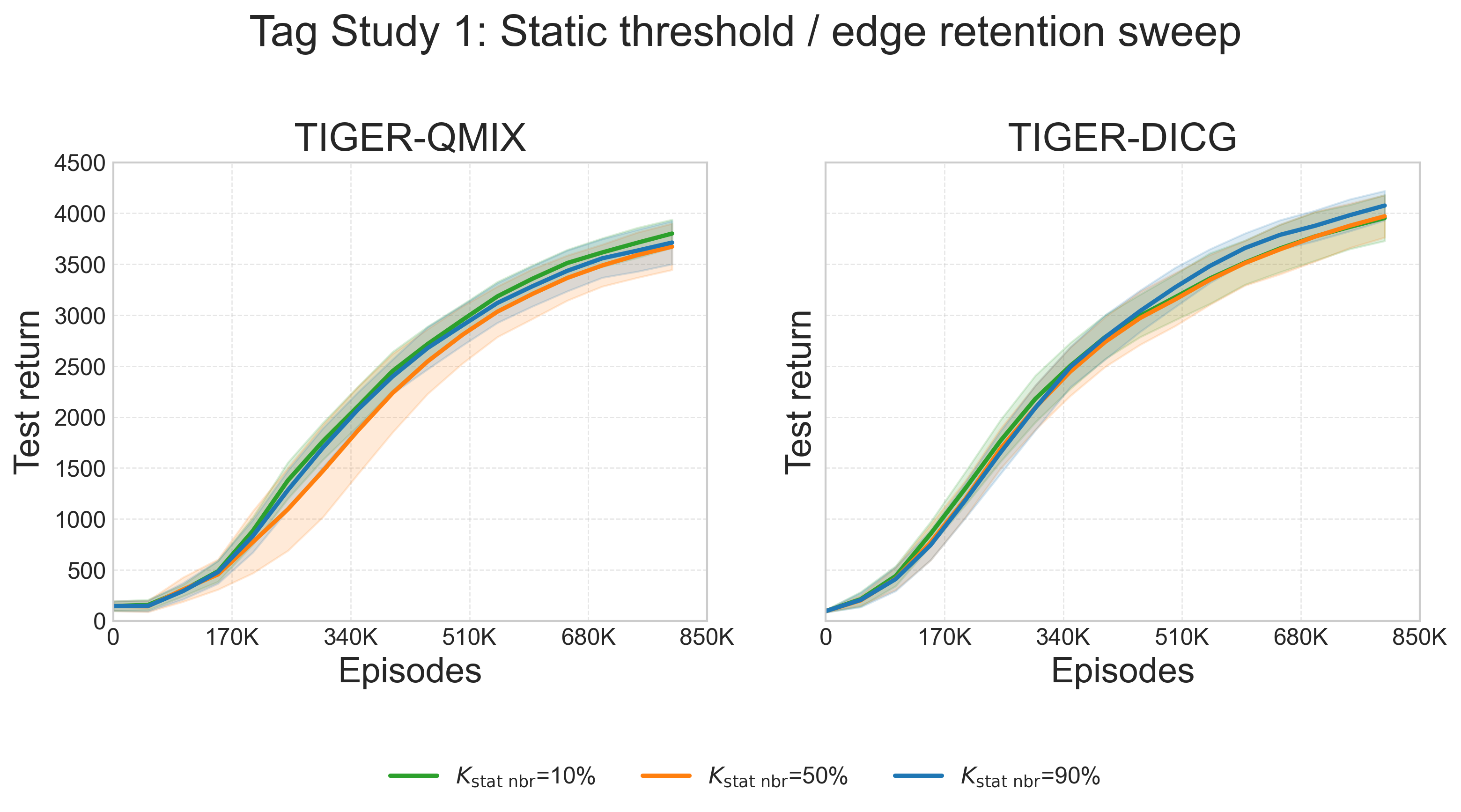}
    \\[-1ex](\alph{subfig})
  \end{minipage}%
  \hfill
  \begin{minipage}{.32\textwidth}
    \centering
    \refstepcounter{subfig}\label{fig:tag_study2}
    \includegraphics[width=\linewidth]{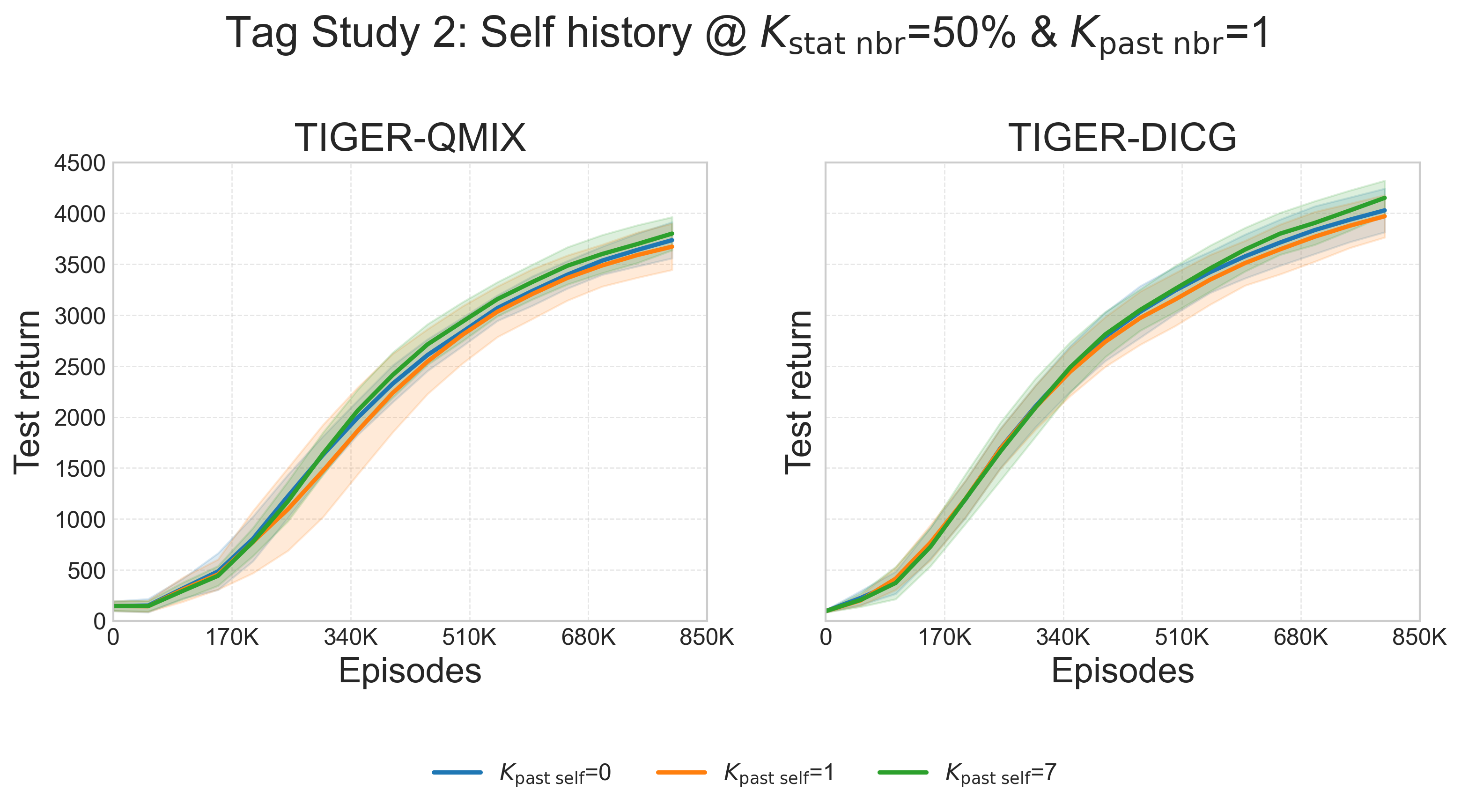}
    \\[-1ex](\alph{subfig})
  \end{minipage}%
  \hfill
  \begin{minipage}{.32\textwidth}
    \centering
    \refstepcounter{subfig}\label{fig:tag_study3}
    \includegraphics[width=\linewidth]{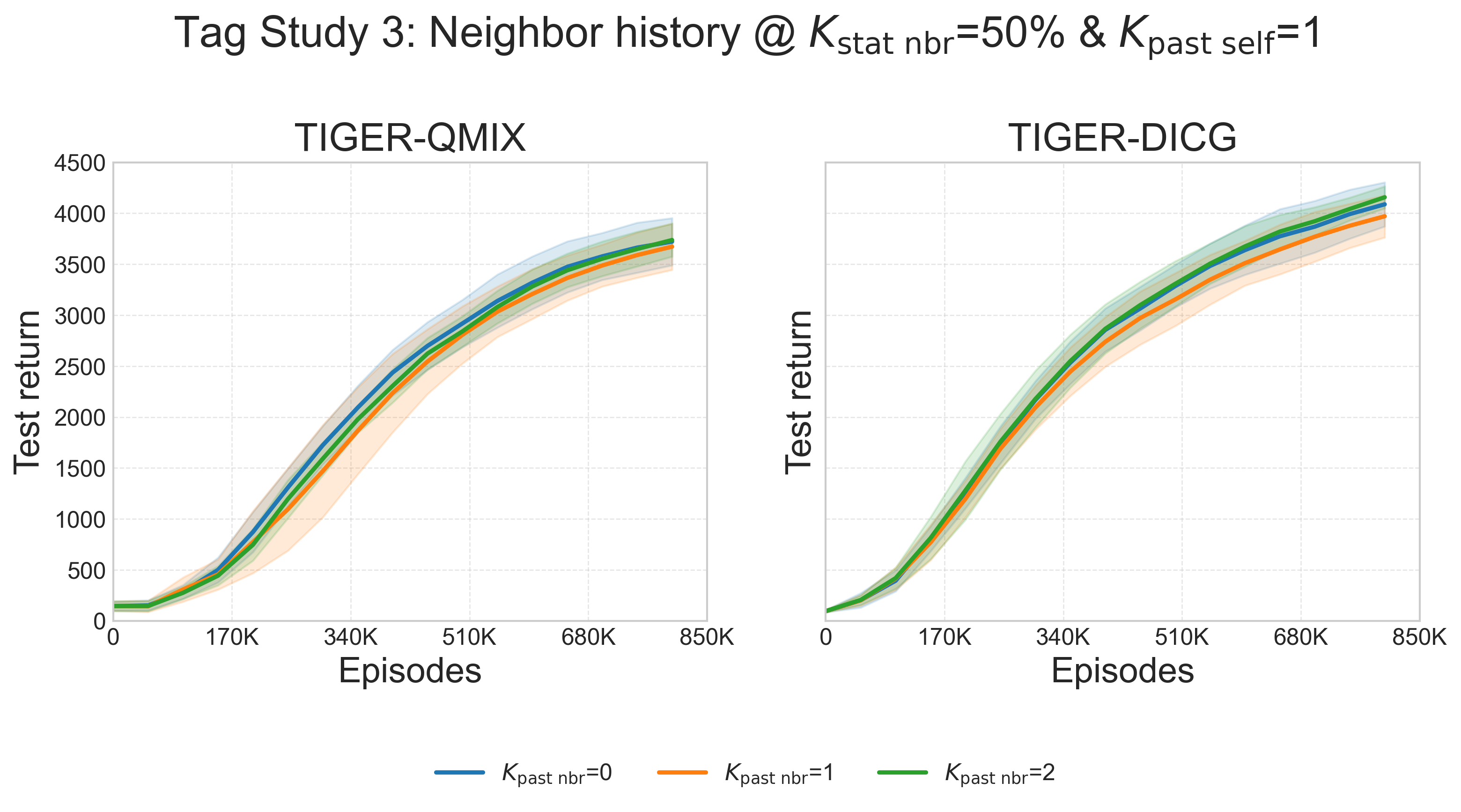}
    \\[-1ex](\alph{subfig})
  \end{minipage}
\caption{\textbf{Ablation studies on \algo.}
(a) Static threshold sweep showing test win rate across neighbor ratios ($K_{\mathrm{stat\_nbr}}$). 
(b) Self-history sweep varying temporal depth ($K_{\mathrm{past\_self}}$). 
(c) Neighbor-history sweep varying neighbor temporal depth ($K_{\mathrm{past\_nbr}}$)}
  \label{fig:tag-ablations13}
\end{figure}

\begin{figure}[!t]
  \centering
  \setcounter{subfig}{0}

  \begin{minipage}{0.48\textwidth}
    \centering
    \refstepcounter{subfig}\label{fig:tag_study4-qmix}
    \includegraphics[width=\linewidth]{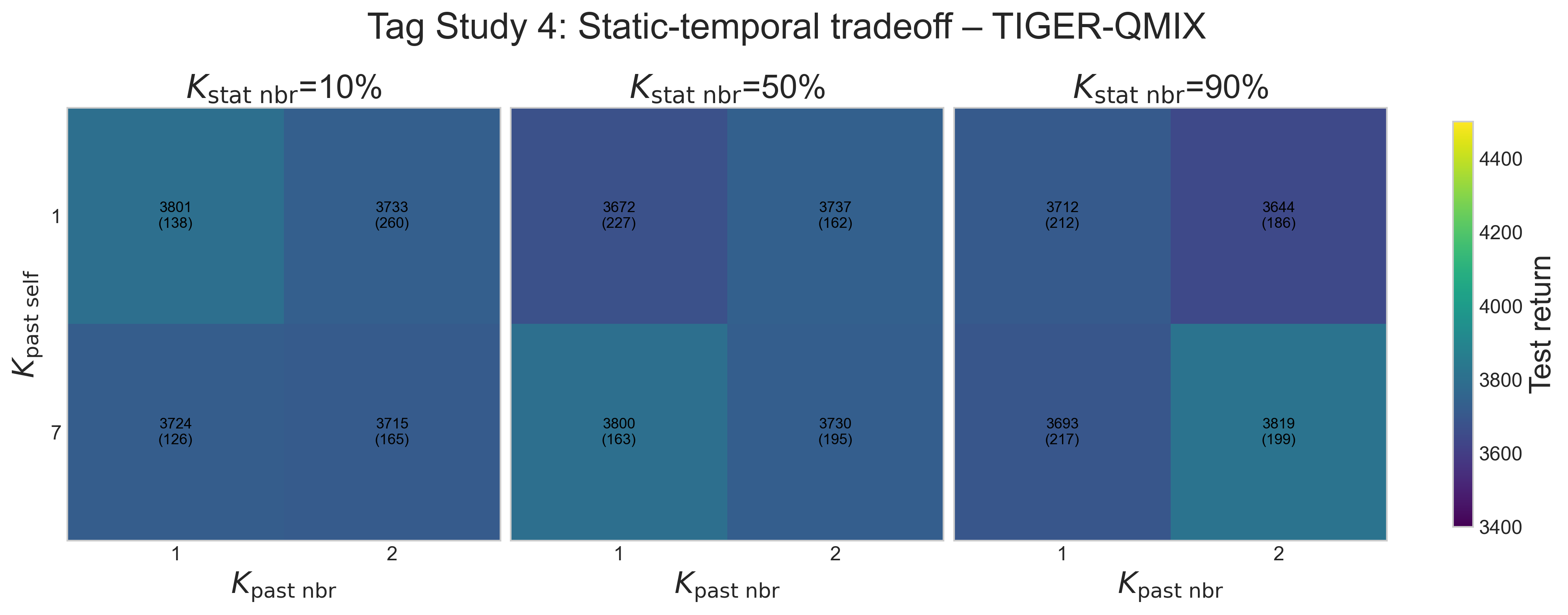} 
    \\[-1ex](\alph{subfig})
  \end{minipage}%
  \hfill
  \begin{minipage}{0.48\textwidth}
    \centering
    \refstepcounter{subfig}\label{fig:tag_study4-dicg}
    \includegraphics[width=\linewidth]{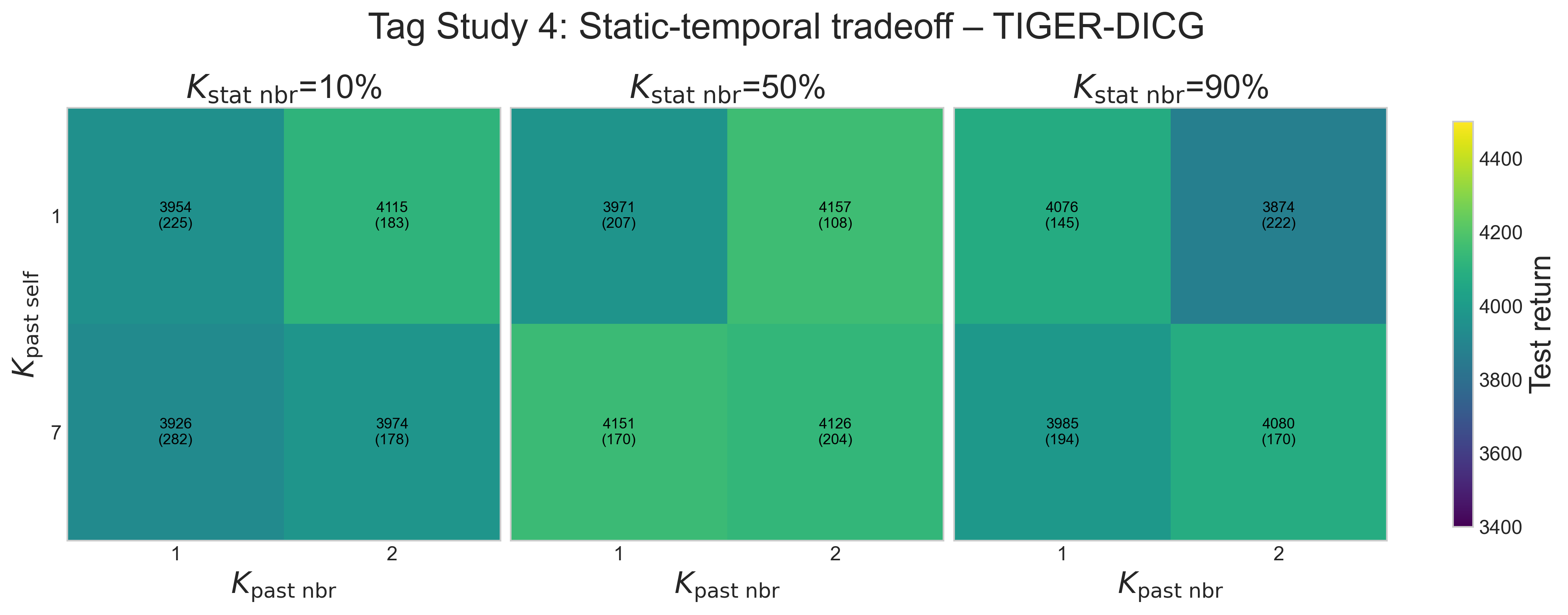} 
    \\[-1ex](\alph{subfig})
  \end{minipage}
\caption{\textbf{Static–temporal tradeoff in \algo.}
(a) \algo-MIX peaks at moderate static connectivity (50\%) and shallow temporal depth (one-hop)
(b) \algo-DICG remains robust across most settings, with 50\% static threshold offering the best balance.}
  \label{fig:tag-study4}
\end{figure}

We conduct ablation studies on the Tag environment following the same procedure as in Gather (see Section~\ref{sec:ablations}) to verify the consistency of \algo’s design choices under distinct coordination dynamics. Unlike Gather, which features delayed rewards and sparse interactions, Tag involves continuous spatial movement, dense encounters, and frequent feedback. We speculate that due to these differences in task structure, configuration, and metrics, the performance variations across \algo's design parameters are less pronounced than in Gather, and definitive conclusions are therefore harder to establish. We also note here that sharp trends are not expected in this setting. The three parameters co-influence a complex, nonlinear learning process involving stochastic agent interactions, attention-driven aggregation, and partial observability. As a result, small changes in temporal depth or connectivity can lead to competing effects, enhanced information flow versus increased variance, which often average out across runs. Nonetheless, some general patterns emerge across both mixers and graph-based \algo-variants in Tag:  
\begin{enumerate}[label=(\roman*),noitemsep,nolistsep,leftmargin=*] 
\item performance remains relatively stable across most configurations, with test returns ranging between  $\sim$3600-3800 for \algo-MIX and $\sim$3900-4200 for \algo-DICG. 
\item moderate static connectivity ($K_{\mathrm{stat\_nbr}}{=}50\%$) consistently provides the best balance, yielding the highest or near-highest returns in both \algo-MIX and \algo-DICG. 
\item deeper temporal recall generally improves coordination, especially in \algo-DICG: returns rise for $K_{\mathrm{past\_self}}{=}7$ and $K_{\mathrm{past\_nbr}}{=}2$, suggesting that richer temporal context enhances adaptation in continuous multi-agent motion. 
\end{enumerate}
Below we further analyze the effects across each design factor, with results shown in Figs.~\ref{fig:tag-ablations13} and~\ref{fig:tag-study4}. For this, we focus on the final absolute test returns to draw concise qualitative conclusions from the Tag ablation studies. 


\paragraph{Static threshold $K_{\mathrm{stat\_nbr}}$ (Fig.~\ref{fig:tag_study1}).}
As in Gather, we begin by reducing static connectivity to a sparse setting ($K_{\mathrm{stat\_nbr}}{=}10\%$). 
For \algo-MIX, this configuration leads to a mild improvement, suggesting that in Tag’s dense feedback regime, fewer structural links help highlight the more relevant interactions. 
Conversely, \algo-DICG\ attains improved results under denser connectivity ($K_{\mathrm{stat\_nbr}}{=}90\%$), indicating that it benefits from richer static relational structure for coordination. 
The intermediate $K_{\mathrm{stat\_nbr}}{=}50\%$ configuration yields comparably stable but slightly lower returns, implying that extreme settings, either sparser or denser, tend to align better with each method. 
Overall, this study on Tag highlights complementary preferences: \algo-MIX benefits from selective sparsity, while \algo-DICG\ thrives in information-rich static relational settings. 

\paragraph{Self history $K_{\mathrm{past\_self}}$ (Fig.~\ref{fig:tag_study2}).}
We next vary the temporal window capturing each agent’s own past states. 
Short memory settings ($K_{\mathrm{past\_self}}=1$) yield slightly lower returns, indicating that limited recall may restrict agents from identifying motion patterns over time. 
When self-history is removed ($K_{\mathrm{past\_self}}=0$), performance remains stable, suggesting that agents can still coordinate effectively using current observations and neighbor context. 
Extending the temporal horizon ($K_{\mathrm{past\_self}}=\ln(NT)=7$) consistently improves both variants, highlighting that deeper self-histories help smooth individual trajectories and enhance adaptation to continuous movement. 
Compared to Gather, where temporal recall offered modest benefits, Tag shows more noticeable advantages from longer self-memory. 

\paragraph{Neighbor history $K_{\mathrm{past\_nbr}}$ (Fig.~\ref{fig:tag_study3}).}
We next analyze the influence of temporal neighbor information by varying $K_{\mathrm{past\_nbr}}$.  
With a shallow temporal window ($K_{\mathrm{past\_nbr}}=1$), performance slightly declines, indicating that limited relational memory may not fully capture the evolving positions and interactions of nearby agents.  
When neighbor history is omitted ($K_{\mathrm{past\_nbr}}=0$), learning remains stable, showing that agents can still coordinate effectively based on immediate context.  
Extending the look-back window to a deeper setting ($K_{\mathrm{past\_nbr}}=2$) consistently enhances performance for both \algo-MIX and \algo-DICG, demonstrating the benefit of richer temporal neighbor information.  
Compared to Gather, this effect is more pronounced in Tag, where fast, spatially dynamic interactions make relational history especially valuable for sustained coordination. 

\paragraph{Static-temporal tradeoff (Fig.~\ref{fig:tag-study4}).}
Finally, we examine the joint influence of static connectivity and temporal depth. 
For \algo-MIX, returns remain largely consistent across settings, showing limited sensitivity to either factor under Tag’s dense feedback. 
In contrast, \algo-DICG\ performs best at moderate static connectivity ($K_{\mathrm{stat\_nbr}}{=}50\%$) and deeper history windows ($K_{\mathrm{past\_nbr}}{=}2$, $K_{\mathrm{past\_self}}{=}7$), indicating that richer temporal context supports its relational reasoning. 
Overall, Tag favors models with deeper temporal recall, though the performance differences are smaller than in Gather, likely due to its denser and more continuously interactive feedback. 

\paragraph{Concluding remarks.}
In summary, these ablation experiments provide some useful insights into the sensitivity of \algo\ to its structural and temporal design parameters.  
While the relative differences between configurations are modest, \algo\ consistently outperforms all non-temporal baselines, reaffirming that temporal reasoning drives its performance gains.  
The observed variations across static connectivity, self-history, and neighbor-history further indicate that each component contributes to coordination quality, confirming that the temporal module captures evolving relational cues rather than acting as a redundant extension.  
Given the inherent complexity of multi-agent coordination, where learning dynamics, environmental stochasticity, and intertwined dependencies are difficult to isolate, these empirical findings invite a deeper theoretical or analytical discussion of temporal graph learning in MARL, which we leave as a promising direction for future work. 

\section{Implementation details} 
\label{app:implementation}

\subsection{More details on MARL baselines} 

\begin{itemize}[noitemsep,leftmargin=*]

\item \textbf{VDN}~\citep{sunehag2018value}:  
VDNs decompose the joint Q-function into the sum of per-agent Q-values, enabling decentralized policies. VDN serves as a foundational baseline without explicit modeling of inter-agent dependencies.  

\item \textbf{QMIX}~\citep{rashid2018qmix}:  
QMIX extends VDN by using a mixing network to combine individual agent Q-values under a monotonicity constraint, facilitating more expressive joint policies while maintaining decentralized execution.  

\item \textbf{QGNN}~\citep{kortvelesy2022qgnn}:  
Graph Neural Network-based Q-value factorization employs message passing to encode agent interactions explicitly. We employ QGNN configurations based on standard implementations provided in the original paper, adjusting the parameters minimally to ensure compatibility with our scenarios.  

\item \textbf{GraphMix}~\citep{naderializadeh2020graph}:  
GraphMix extends the QMIX architecture by integrating static graph neural networks to improve relational reasoning among agents. We utilize the standard GraphMix settings with the default fully connected configuration as suggested in the corresponding literature.  

\item \textbf{DICG}~\citep{li2021deepimplicitcoordinationgraphs}:  
Deep Implicit Coordination Graphs utilize attention mechanisms to dynamically generate weighted, fully connected coordination graphs among agents. Information is passed using learned attention weights. We strictly follow the original DICG methodology, employing QMIX as the underlying CTDE framework.  

\item \textbf{CASEC}~\citep{wang2022contextaware}:  
Context-Aware Sparse Deep Coordination Graphs selectively prune edges from fully connected coordination graphs based on variance estimation of payoffs. We configure CASEC using the variance-based edge pruning (as defined in the original paper) and set the sparsity loss weighting parameter $\lambda_{sparse} = 0.3$.  

\item \textbf{GACG}~\citep{duan2024groupawarecoordinationgraphmultiagent}:  
Group-Aware Coordination Graphs learn dynamic groupings among agents to capture higher-order relationships. We adopt recommended hyperparameters from the original implementation to effectively model agent group dependencies.  

\item \textbf{LTSCG}~\citep{duan2024inferring}:  
Latent Temporal Sparse Coordination Graphs infer sparse temporal graphs by combining edge probability learning with historical interaction encoding. We use the official codebase and training protocols.  

\end{itemize} 

\subsection{MARL training setup}

All agents share the same network configuration to ensure a fair comparison across baselines. Each agent uses a two-layer GRU (64 hidden units) to encode its local observation and produce an individual $Q$-vector. Mixer-based algorithms employ a two-layer hypernetwork with ReLU activations to compute aggregation weights, whereas graph-based learners treat agent embeddings as nodes and perform one round of message passing with 32-dimensional channels. Training settings are kept consistent across all methods to isolate the effect of temporal graph reasoning. $\epsilon$-greedy exploration is linearly annealed from $1.0$ to $0.05$ over the first $200$K timesteps. Optimization uses Adam ($5\times10^{-4}$ learning rate), discount factor $\gamma=0.99$, and TD($\lambda$) with $\lambda=0.8$. The replay buffer stores the most recent 5000 episodes, from which 32 mini-batches are sampled per update. Target networks are synchronized every 200 learner steps, and gradients are clipped to a maximum norm of 10 to ensure stability. 

\section{Discussion on the computational complexity} 
\label{app:complexity}

\paragraph{Size of Temporal Graph.} 
Let \(N\) denote the number of agents and \(T\) the episode horizon. 
If the interaction graph is time-unrolled, it yields \(N T\) nodes in total. 
The number of edges decomposes into three main components:  
(i) \textbf{Static neighbors:} within each timestep, every agent can connect to up to \(N{-}1\) others, resulting in \(\mathcal{O}(N^2)\) edges per step;  
(ii) \textbf{Self-history:} each agent connects to its own past, producing \(\mathcal{O}(N T)\) edges at time \(t\) and \(\mathcal{O}(N T^2)\) over an episode;  
(iii) \textbf{Past neighbors:} agents connect to their neighbors’ past states, yielding \(\mathcal{O}(N^2 T)\) edges at time \(t\) and \(\mathcal{O}(N^2 T^2)\) over the full episode.  
Hence, the total temporal graph complexity across the episode is dominated by 
\[
E_{\text{total}} = \mathcal{O}(N^2 T^2),
\]
indicating quadratic growth in both the number of agents and the horizon length.

\paragraph{Examples.} For Gather with \(N{=}5\) agents and \(T{=}8\) timesteps, the temporal graph contains \(N T = 40\) nodes. 
The static edges are \(\frac{5\times4}{2} \times 8 = 80\) across the episode, while self-history and past-neighbor edges accumulate to approximately \(5 \times \frac{8\times7}{2} = 140\) and \(\frac{5\times4}{2} \times ({8\times7}) = 560\) respectively.  
Simailarly, for Tag, with \(N{=}13\) and \(T{=}100\), there are \(1{,}300\) nodes, \(78\) static edges, and roughly \(64{,}350\) and \(772{,}200\) edges from self- and neighbor-history respectively.  
This demonstrates how long horizons and dense interactions can produce rapid edge growth when temporal graphs are reconstructed at each timestep.

\paragraph{Controlling graph size.} 
As shown in the above examples, even with relatively small agent counts and horizons, temporal graph construction can quickly grow in scale. \algo\ introduces control parameters that directly mitigate this blow-up: 
\(K_{\mathrm{stat\_nbr}}\) limits static neighbor connectivity, 
\(K_{\mathrm{past\_self}}\) bounds each agent’s self-history window, and 
\(K_{\mathrm{past\_nbr}}\) restricts how many neighbor histories are considered. 
These choices cap the effective complexity to a tractable subset of the full temporal graph, reducing redundant edges while retaining relevant relational structure. 
The scalability and flexibility of \algo’s temporal graph construction are particularly relevant to real-world MARL domains where agent coordination evolves over time. 
In autonomous driving or drone swarm control, where decisions unfold over long horizons and coordination structures change rapidly, deeper temporal recall (\(K_{\mathrm{past\_self}}\), \(K_{\mathrm{past\_nbr}}\)) allows agents to leverage extended behavioral history and adapt to dynamic formations. 
In contrast, short-horizon or event-triggered domains such as multi-robot assembly or distributed sensing can benefit from lightweight configurations with shallow temporal windows and sparse connectivity (\(K_{\mathrm{stat\_nbr}}\)), reducing computation without compromising coordination fidelity. 
Thus, the ability to regulate graph density and temporal depth enables \algo\ to scale computationally across domains with different coordination requirements, maintaining efficiency in resource-constrained systems while preserving expressivity in complex environments.

\end{document}